\documentclass{article} 
\usepackage{iclr2026_conference,times}

\usepackage{amsmath,amsfonts,bm}

\def\eqref#1{equation~\ref{#1}}

\def\1{\bm{1}}

\DeclareMathAlphabet{\mathsfit}{\encodingdefault}{\sfdefault}{m}{sl}
\SetMathAlphabet{\mathsfit}{bold}{\encodingdefault}{\sfdefault}{bx}{n}

\usepackage{hyperref}
\usepackage{url}

\usepackage{amsmath,amssymb,amsfonts}
\usepackage{algorithmic}
\usepackage{graphicx}
\usepackage{textcomp}
\usepackage{subcaption}
\usepackage{multirow}
\usepackage{caption} 
\usepackage{booktabs}
\usepackage[table]{xcolor} 
\usepackage{xcolor}
\definecolor{mygreen}{RGB}{0,128,0}
\definecolor{myred}{RGB}{255,0,0} 
\definecolor{lightgray}{gray}{0.92}
\definecolor{lightyellow}{rgb}{1,1,0.85}

\title{Informed Routing in LLMs: Smarter Token-Level Computation for Faster Inference}

\author{%
 \textbf{Chao Han\textsuperscript{1}},\;
 \textbf{Yijuan Liang\textsuperscript{1,2}},\;
 \textbf{Zihao Xuan\textsuperscript{3}},\;
 \textbf{Daokuan Wu\textsuperscript{1}},\;
 \textbf{Wei Zhang\textsuperscript{1}},\;
 \textbf{Xiaoyu Shen\textsuperscript{1\thanks{Corresponding Author}}}\\[4pt]
 \textsuperscript{1}Institute of Digital Twin, Eastern Institute of Technology, Ningbo\\
 \textsuperscript{2}University of Science and Technology of China\\
 \textsuperscript{3}The Hong Kong University of Science and Technology
\\
 \texttt{chan@eitech.edu.cn\quad xyshen@eitech.edu.cn}\\[3pt]
}

\iclrfinalcopy 
\begin{document}

\maketitle

\begin{abstract}
The deployment of large language models (LLMs) in real-world applications is increasingly limited by their high inference cost. While recent advances in dynamic token-level computation allocation attempt to improve efficiency by selectively activating model components per token, existing methods rely on greedy routing—a myopic execute-or-skip mechanism that often leads to irreversible information loss and suboptimal token selection.
This paper introduces informed routing, a new paradigm that proactively addresses these issues. The key insight is to assess not only a token’s immediate importance but also its recoverability, i.e., how well its transformation can be approximated. To this end, we propose the Lightweight Feature Forecaster (LFF), a small predictive module that estimates a unit’s output before routing decisions are made. This enables a flexible execute-or-approximate policy that preserves model fidelity while drastically reducing computation.
Extensive experiments on both language modeling and reasoning tasks show that informed routing achieves state-of-the-art efficiency–performance trade-offs across multiple sparsity levels. Notably, even without final LoRA fine-tuning, our method matches or surpasses strong baselines that require full fine-tuning, all while reducing training time by over 50\%. The code is available in the GitHub repository: \url{https://github.com/EIT-NLP/informed-routing}.
\end{abstract}

\section{Introduction}
The emergence of large language models (LLMs) has catalyzed breakthroughs across diverse industries~\citep{su2022welm,openai2024gpt4technicalreport, rozière2024codellamaopenfoundation,cai2025large,Zheng2025}. Scaling laws have established computational requirements as a primary bottleneck in the development and deployment of LLMs~\citep{kaplan2020scalinglawsneurallanguage,su2024unraveling}. Therefore, reducing this computational overhead has become a key research objective.

Early work primarily focused on \textbf{static pruning} methods, which permanently remove a fixed subset of parameters or components from the model~\citep{han2016deep, ma2023llm,xu2025rethinkingvisualtokenreduction}. While effective for compression, these approaches fail to exploit the varying importance of tokens during inference. More recently, the observation of diverse token criticality has motivated a shift toward \textbf{dynamic computation allocation (DCA)}~\citep{raposo2024mixture,jiang2024d,roe,shin2025orthorank}, where different tokens undergo different amounts of computation. DCA partitions the model into computational units—ranging from coarse-grained layers to finer-grained sub-layer components (e.g., self-attention blocks and feed-foward network blocks within a single layer~\citep{zhao2025skipgpt}), each equipped with a router. These routers, typically small MLPs, are trained post-hoc to decide whether to execute or skip a unit for each token. In practice, important tokens are routed through most of the model’s parameters, while less important ones can skip substantial computation. This flexibility mirrors human language processing, where critical words are analyzed in depth while less informative ones receive only shallow processing.

\begin{figure}[tbp]
    \centering
    \begin{subfigure}[b]{0.48\textwidth}
        \centering
        \includegraphics[width=\textwidth]{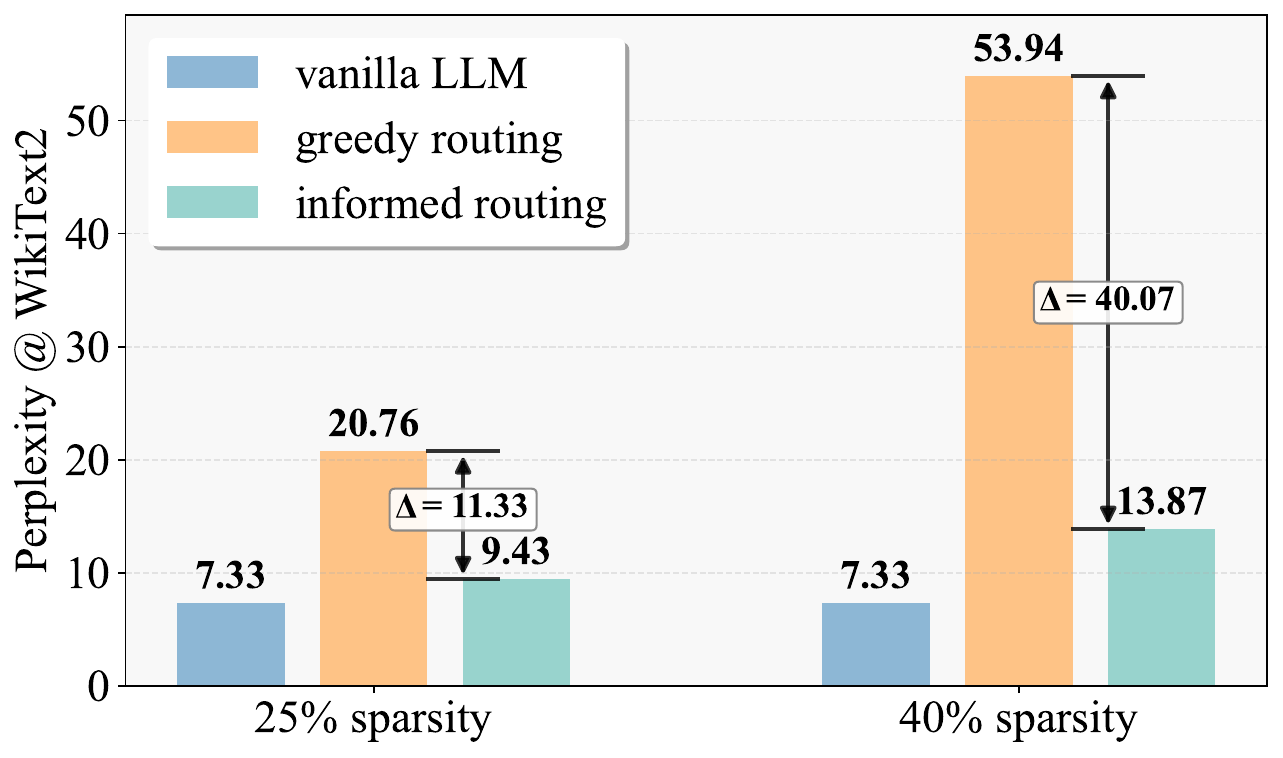}
        \caption{Perplexity After Router Training}
        \label{fig:sub1}
    \end{subfigure}
    \hfill 
    \begin{subfigure}[b]{0.48\textwidth}
        \centering
        \includegraphics[width=\textwidth]{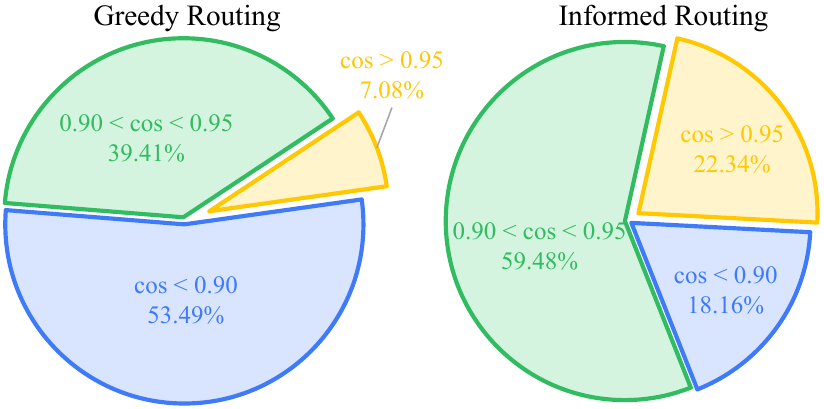}
        \caption{Cosine Similarity Distribution of Features}
        \label{fig:sub2}
    \end{subfigure}
    
    \caption{\small{The limitation of greedy routing and the promise of informed routing. Under the same sparity ratio, informed routing (a) reduces the perplexity and (b) increases feature similarity.}}
    \label{fig:main}
\end{figure}

However, existing DCA methods are constrained by a paradigm we term \textbf{greedy routing}. Routers are trained to make a simple, binary choice: fully execute a computational unit or skip it entirely. Performance recovery is then attempted via lightweight fine-tuning (e.g., LoRA~\citep{hu2022lora}).~\footnote{Some works have also explored jointly training the router and the recovery module~\citep{jiang2024d}, however, empirical evidence suggests that this can lead to performance degradation \citep{zhao2025skipgpt}.}. The decision to skip is based on minimizing the \textit{immediate} performance drop, without considering the long-term consequences. This greedy approach suffers from two fundamental flaws:
\begin{enumerate}
    \item \textbf{The All-or-Nothing Dilemma:} By forcing a rigid execute-or-skip decision, this paradigm offers no middle ground. Skipping a unit causes irreversible information loss, disrupting the model’s internal feature distributions and requiring costly fine-tuning to recover performance. As shown in Figure \ref{fig:sub1}, skipping tokens leads to a significant increase in perplexity, from 7.33 to 20.76 (25\% sparsity) and 53.94 (40\% sparsity). 
    \item \textbf{Short-Sighted Token Selection:} The router’s focus on immediate impact is a poor proxy for true importance. A token that causes a large immediate drop when skipped is not necessarily indispensable; its transformation might be simple and easily recoverable later. Conversely, a token with low immediate impact might be crucial for maintaining subtle, long-range dependencies that are difficult to restore once lost.
\end{enumerate}

To overcome these limitations, we propose a paradigm shift from greedy decision-making to \textbf{informed routing}. Our central idea is to replace the binary execute-or-skip choice with a more nuanced execute-or-approximate decision. We achieve this by equipping each computational unit with a \textbf{Lightweight Feature Forecaster (LFF)}—a small, efficient network trained to mimic the output of its larger counterpart. The router’s task is no longer to guess which tokens can be safely dropped, but to determine which tokens are \textit{predictable}. If a token’s transformation can be accurately approximated by the LFF, it is routed through this efficient path. If the transformation is too complex to be forecasted, the token is processed by the original, powerful unit.

This ``informed'' approach allows the router to learn a policy based on a token's \textbf{recoverability}, not just its immediate impact. For instance, an analysis of feature similarity (Figure \ref{fig:sub2}) reveals that LFF increases the proportion of features that remain highly similar (cosine similarity $> 0.95$), from 7.08\% to 22.34\%. Such significant increase provides concrete evidence for the existence of a substantial fraction of \textit{predictable} tokens—tokens whose transformations can be accurately approximated by the LFF. Based on this insight, we propose a simple three-stage pipeline: (1) train LFF to approximate their corresponding units, (2) train routers to choose between the original unit and its LFF for each token, and (3) perform optional, lightweight LoRA~\citep{hu2022lora} fine-tuning to polish the final model. Importantly, we show that we could shrink the original router’s intermediate hidden size and reuse the freed capacity to host the LFF, such that \emph{the overall parameter count and computational cost remain identical to standard DCA}, ensuring a fair comparison in both efficiency and resource usage.

Our contributions are as follows:
\begin{enumerate}
    \item We identify the core limitations of the prevailing \textbf{greedy routing} paradigm in DCA, namely its rigid all-or-nothing mechanism and its short-sighted reliance on immediate impact as a routing criterion.
    \item We introduce \textbf{informed routing}, a new paradigm enabled by \textbf{Lightweight Feature Forecaster (LFF)}, which replaces skipping with efficient approximation and allows routing decisions to be based on a token's recoverability.
    \item We demonstrate through extensive experiments that our method achieves state-of-the-art efficiency, significantly reducing training overhead and improving final performance. Our analysis further reveals that self-attention modules are highly ``predictable'', making them prime candidates for approximation.
\end{enumerate}

\section{Related Works}
\paragraph{Static Pruning}
Broadly speaking, static pruning techniques related to our approach fall into two main categories: token pruning and parameter pruning.
\textbf{Static Token Pruning} methods identify and remove tokens deemed redundant, and pruned tokens bypass subsequent transformer layers, which are widely used in Vision-Language Models(VLMs). SpecVLM\citep{ji2025SpecVLM} introduces a 'verifier' model to estimate the importance of video tokens.  VisionDrop\citep{xu2025rethinkingvisualtokenreduction} identifies token importance via intra-modal attention. In contrast, our dynamic approach achieves flexibility along \textit{model depth}, allowing each token to undergo full computation only in the layers where it is most needed, thereby preserving information while adaptively saving computation.
\textbf{Static Parameter Pruning} techniques permanently remove fixed structural components(e.g., layers or neurons), resulting in a uniformly smaller model. SliceGPT\citep{ashkboos2024slicegpt} employs Principal Component Analysis (PCA) on the orthogonally transformed parameters, followed by the removal of entire rows/columns.  Shortened-llama\citep{kim2024shortened} demonstrates that depth pruning is more efficient than width for LLM inference. ShortGPT\citep{men2024shortgptlayerslargelanguage} proposes Block Influence(BI) to quantitatively estimate the importance of layers in large language models, and subsequently prunes the less important layers.  LLM-Streamline\citep{ICLR2025_4b00a351}  removes consecutive layers and then replaces them with a smaller model.  Parameter reduction in capacity applies inflexibly to all tokens, regardless of their importance. Conversely, the proposed dynamic method provides flexibility across \textit{input token sequence}, i.e. for a given parameter structure, only a subset of tokens utilizes it while others bypass it via a lightweight path (LFF), enabling a more granular and input-adaptive efficiency.

\paragraph{Dynamic Computation Allocation}
Dynamic computation allocation methods leverage the observation that linguistic representations evolve at varying paces across tokens. Central to these techniques is the \textit{router} mechanism—a lightweight classifier that dynamically assigns computational paths to tokens. The router acts as a per-token binary classifier at each computation unit. For every token representation, it first computes skip/keep probabilities using a multilayer perceptron. The execution path is then determined via argmax sampling: if ``skip'' is selected, the token bypasses the unit and remains unchanged; if ``keep'' is chosen, it undergoes transformation within the unit. This gating mechanism results in dynamic, token-wise computation graphs where inactive tokens propagate directly to the next layer without being processed.
Contemporary approaches to dynamic computation allocation exhibit notable variations in sparsity control and computational granularity. Mixture-of-Depths (MoD) \citep{raposo2024mixture}, enforces a fixed sparsity ratio per layer block, which limits its ability to adapt to input-specific redundancy patterns and ultimately constrains its efficiency. In contrast, subsequent work such as D-LLM  \citep{jiang2024d} introduces global adaptive sparsity, dynamically allocating computation across layers in response to input characteristics. Building on this, SkipGPT \citep{zhao2025skipgpt} further refines the granularity by decoupling attention and MLP operations within each layer. It employs separate routers to independently skip each sub-module, enabling more flexible computation paths while maintaining globally optimized sparsity. However, all of these are built with greedy routing.

\paragraph{Error Compensation}
Prior works have explored error compensation to mitigate compression-induced accuracy drops. RECAP \cite{lee2025recap} transfers the statistics of pruned channels to adjacent weights. PRUNE\&COMP \cite{chen2025prune} rescales remaining weights offline to compensate for the magnitude gap after layer removal. Olica \cite{he2025olica} introduces a linear mapping for low-rank compensation in FFNs.  ROE \cite{roe} also identifies the feature gap issue caused by router skipping in multimodal QA tasks, and implements lightweight network adaptation (termed as adapter) at the instance level.  While these methods perform \textit{weight-wise} , \textit{channel-wise} or \textit{instance-wise} compensation, our approach operates in a \textit{token-wise} manner. We dynamically route tokens to either the original model or a LFF, enabling finer-grained and adaptive error recovery. This allows critical tokens to retain full precision while approximating redundant ones, achieving more flexible accuracy-efficiency trade-offs. Additionally, compared with the global alignment of ROE's adapter, our method adopts a local alignment approach for LFF initialization. Specifically, gradients do not need to pass through LLM—this makes the training time of LFF almost negligible compared to the subsequent router/LoRA training, i.e. 5 minutes vs. 3 hours. This aspect constitutes a major contribution of this paper: under 25\%/40\% sparsity, the proposed informed routing can match or even outperform the performance of greedy routing while saving half of the training time.

\section{Methodology}
\label{sec:method}
\begin{figure*}[htbp]
  \centering
  \includegraphics[width=1.0\linewidth]{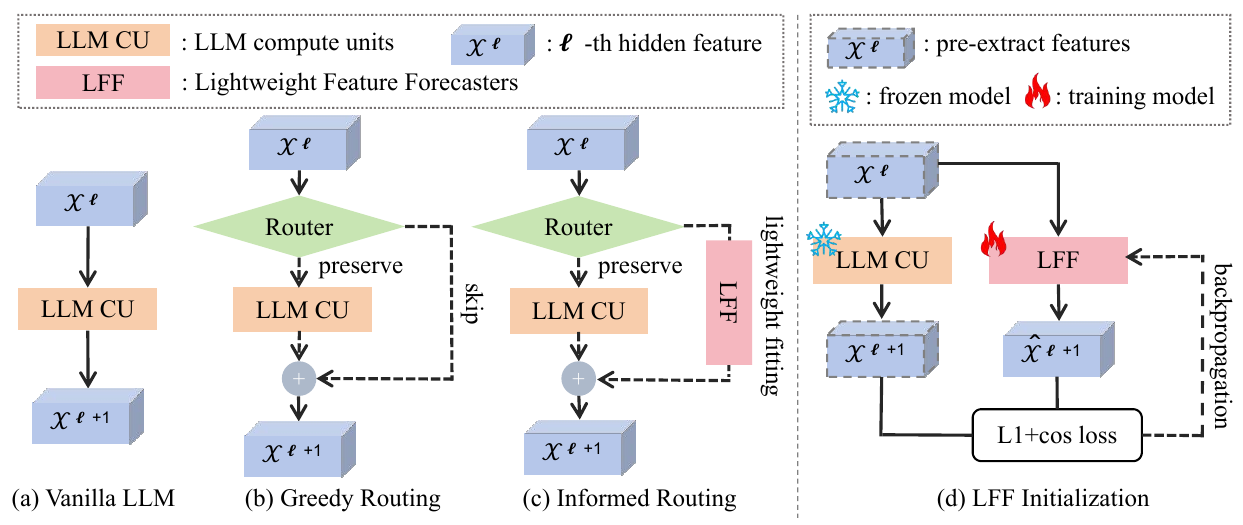}
  \caption{\small{(a), (b), and (c) present the architectural comparison diagrams of the vanilla LLM, greedy routing, and our proposed informed routing paradigm. (d) illustrates how the LFF initialization is performed.}} 
  \label{fig:arch} 
\end{figure*}

In this section, we introduce \textbf{informed routing}, a framework that replaces rigid skip decisions with efficient approximations. Figure~\ref{fig:arch} provides an overview of the framework, and we now detail the architecture, the LFF design, and the training procedure.

\subsection{Preliminaries and Notation}
Consider a transformer-based LLM with $L$ layers. For layer $\ell \in \{1,\dots,L\}$, let $\mathbf{X}^{\ell} \in \mathbb{R}^{N \times d}$ denote the input token embeddings, where $N$ is sequence length and $d$ is the hidden dimension. Each transformer layer $\ell$ processes the input through two components, i.e. self-attention and feed-forward network(FFN), with pre-normalization and residual connections:

\begin{align}
\mathbf{X}^{\ell}_{\text{att}} &= \mathcal{A}^{\ell}\left(\text{Norm}(\mathbf{X}^{\ell})\right) + \mathbf{X}^{\ell} \quad \text{(Self-attention module)} \\
\mathbf{X}^{\ell+1} &= \mathcal{F}^{\ell}\left(\text{Norm}(\mathbf{X}^{\ell}_{\text{att}})\right) + \mathbf{X}^{\ell}_{\text{att}} \quad \text{(FFN module)}
\end{align}

where:
$\mathcal{A}^{\ell}$: Multi-head self-attention at layer $\ell$,
$\mathcal{F}^{\ell}$: Feed-forward network at layer $\ell$,
$\mathbf{X}^{\ell}_{\text{att}}$: Intermediate representation after attention,
$\mathbf{X}^{\ell+1}$: Output embeddings serving as input to layer $\ell+1$.

Following SkipGPT's granularity, we decompose each transformer layer $\ell$ into two \textit{computational units}: $\mathcal{U}^{\ell,\text{SA}}$ (self-attention) and $\mathcal{U}^{\ell,\text{FFN}}$ (feed-forward network). For each unit, a lightweight \textit{router} $\mathcal{R}^{\ell,k}: \mathbb{R}^d \to \mathbb{R}^2$ (where $k \in \{\text{SA}, \text{FFN}\}$) makes token-wise pruning decisions. The router is implemented as a two-layer MLP with bottleneck dimension, i.e $\mathbb{R}^d \to \mathbb{R}^{\lfloor d/4 \rfloor} \to \mathbb{R}^2$.

The router outputs decision logits for each token $\mathbf{x}_i^{\ell,k}$:
\begin{equation}
\mathbf{r}_i^{\ell,k} = \mathcal{R}^{\ell,k}(\mathbf{x}_i^{\ell,k}) \in \mathbb{R}^2
\end{equation}
with routing probabilities obtained via softmax:
\begin{equation}
\label{eq:p}
p_i^{\ell,k} = \sigma(\mathbf{r}_i^{\ell,k})_1 = \frac{\exp(\mathbf{r}_i^{\ell,k}[1])}{\sum_{c=0}^1 \exp(\mathbf{r}_i^{\ell,k}[c])}
\end{equation}
where class $c=1$ indicates \textit{preserving precision through original LLM compute unit} and $c=0$ indicates \textit{lightweight fitting via LFF branch}. Figure \ref{fig:arch}(c) illustrates this computational flow.

During the forward pass, a hard binary mask $\mathbf{p}^{\ell,k}$ is sampled by applying the $\arg\max$ operation to Gumbel-Softmax logits, producing discrete ${0,1}$ values. In the backward pass, the gradient is estimated using a continuous softmax approximation with temperature $\tau$, enabling differentiable training. The temperature is annealed linearly from $\tau=5.0$ to $\tau=1.0$ to sharpen the distribution over time. Modern frameworks such as PyTorch provide built-in functions like $F.gumbel\_softmax$, which facilitates end-to-end training of discrete latent variable models.

The training objective minimizes computation while preserving performance by enforcing a target relative sparsity $S_{\text{target}}$ (e.g., 50\%). The global computation fraction is:
\begin{equation}
\label{eq:rho}
\rho = \frac{1}{2LN} \sum_{\ell=1}^L \sum_{k \in \{\text{SA},\text{FFN}\}} \|\mathbf{p}^{\ell,k}\|_0
\end{equation}
where $\|\mathbf{p}^{\ell,k}\|_0 = \sum_i p_i^{\ell,k}$, and we regulate $\rho$ toward $S_{\text{target}}$ during training (Section \ref{subsec:training}).

\subsection{Lightweight Feature Forecaster}
\label{subsec:forecasters}
The core innovation of our paper is the \textbf{lightweight feature forecaster} $\mathcal{F}^{\ell,k}: \mathbb{R}^d \to \mathbb{R}^d$ that approximates the input-output mapping of computational unit $\mathcal{U}^{\ell,k}$ \textit{before} routing decisions. This architectural shift transitions the routing paradigm from reactive recovery to proactive preservation. For efficiency, $\mathcal{F}^{\ell,k}$ uses a bottleneck architecture: 
\begin{equation}
\mathcal{F}^{\ell,k}(\mathbf{x}) = \mathbf{W}_2^{\ell,k} \cdot \left(\mathbf{W}_1^{\ell,k} \mathbf{x} + \mathbf{b}_1^{\ell,k}\right) + \mathbf{b}_2^{\ell,k}
\end{equation}
where $\mathbf{W}_1^{\ell,k} \in \mathbb{R}^{r \times d}$, $\mathbf{W}_2^{\ell,k} \in \mathbb{R}^{d \times r}$ with $r \ll d$ (e.g., $r=100$ with $d=4096$ for Llama3.1-8B \citep{grattafiori2024llama}). This yields minimal parameters: $ 4096 \times 100 + 100 \times 4096  \approx 0.82\text{M}$ ($0.02\%$ of $\mathcal{U}^{\text{FFN}}$).

$\mathcal{F}^{\ell,k}$ predicts $\mathcal{U}^{\ell,k}$'s normalized output $\mathbf{z}_i^{\ell,k} \triangleq \text{Norm}\left(\mathcal{U}^{\ell,k}(\mathbf{x}_i^{\ell,k})\right)$ using $cosine\ similarity\ loss$ and $L_1\ loss$. Obviously, when the LFF outputs all zeros (i.e., when all its weights are zero), our method degenerates to greedy routing. 

\subsection{Three-Stage Optimization}
\label{subsec:training}

\noindent\textbf{Stage 1: LFF Initialization.} 
We commence by training the feature forecasters $\mathcal{F}^{\ell,k}$ to approximate the functional mapping of each computational unit $\mathcal{U}^{\ell,k}$. During this phase, the base LLM parameters remain \textit{frozen}, preserving the original feature distributions. For each unit $(\ell,k)$, we minimize the forecasting loss $\mathcal{L}_{\text{fit}}^{\ell,k}$ using feature pairs $\{(\mathbf{x}_i^{\ell,k}, \mathbf{z}_i^{\ell,k})\}_{i=1}^N$ extracted from a random-selected subset (2,000 samples) of the training corpus. 

This decoupled training paradigm (as shown in Figure~\ref{fig:arch}(d)) admits two significant advantages:
\begin{enumerate}
    \item \textit{Architectural Independence}: Each $\mathcal{F}^{\ell,k}$ learns a \textit{local approximation} of $\mathcal{U}^{\ell,k}$ without gradient propagation between computational units. This isolation eliminates inter-unit dependencies, enabling:
    
    \item \textit{Massive Parallelization}: Forecasters across all $L$ layers and $k \in \{\text{SA}, \text{FFN}\}$ can be trained concurrently via:
    \[
    \underset{\theta_{\mathcal{F}^{\ell,k}}}{\text{minimize}} \mathbb{F}_{(\mathbf{x},\mathbf{z}) \sim \mathcal{D}} \left[ \mathcal{L}_{\text{fit}}^{\ell,k} \left( \mathcal{F}^{\ell,k}(\mathbf{x}; \theta), \mathbf{z} \right) \right] \quad \forall (\ell,k)
    \]
    where $\theta$ denotes forecaster parameters and $\mathcal{D}$ the feature dataset, ensuring computational efficiency during this stage.
\end{enumerate}

Feature tensors $\mathbf{X}^{\ell,k}$ and $\mathbf{Z}^{\ell,k}$ can be precomputed offline, circumventing GPU memory bottlenecks associated with full-model activations. For LLaMA3.1-8B (with 64 LFF), this stage completes in less than 5 minutes on a single NVIDIA RTX 6000 Ada GPU (48GB VRAM).

\noindent\textbf{Stage 2: Router Training.}
Jointly train routers $\{\mathcal{R}^{\ell,k}\}$ with LLM and forecasters frozen. Router architecture: 
\begin{equation}
\mathcal{R}^{\ell,k}(\mathbf{x}) = \text{Linear}_{{\lfloor d_1 \rfloor} \to 2}\left(\text{ReLU}\left(\text{Linear}_{d \to {\lfloor d_1 \rfloor} }\left(\mathbf{x}\right)\right)\right)
\end{equation}

Although the LFF is already sufficiently lightweight, to ensure a fair comparison with baseline methods, we compromise on the parameter configuration of the router to match the parameter count of SkipGPT. Specifically, the intermediate dimension ($d_1$) of our router is 200 lower than that of SkipGPT. For instance, in the case of the Llama3.1-8B model, SkipGPT adopts an intermediate dimension of \(4096/4 = 1024\), while we use \(4096/4 - 200 = 824\). As a result, SkipGPT introduces a total of 268.56M parameters (routers), whereas we introduce 268.54M parameters (routers + LFF).

Composite loss integrates:
\begin{align}
\mathcal{L}_{\text{route}} &= \mathcal{L}_{\text{LM}} + \lambda_1 \mathcal{L}_{\text{sparse}}  = \mathcal{L}_{\text{LM}} + || \rho - S_{\text{target}} ||_{1} 
\end{align}
where $\mathcal{L}_{\text{LM}}$ is the language modeling loss, $\rho$ is global compute fraction (Eq.~\ref{eq:rho}) and $\lambda_1=8.0$ balances two objective.

\noindent\textbf{Stage 3: Parameter-Efficient Fine-tuning.}
Inject LoRA adapters into attention projections ($\mathbf{W}_Q,\mathbf{W}_K,\mathbf{W}_V$) and FFN gates: 
\begin{equation}
\mathbf{W} \leftarrow \mathbf{W} + \mathbf{A}\mathbf{B}, \quad \mathbf{A} \in \mathbb{R}^{d \times r_{\text{LoRA}}}, \mathbf{B} \in \mathbb{R}^{r_{\text{LoRA}} \times d}, \; r_{\text{LoRA}}=16
\end{equation}
Minimize $\mathcal{L}_{\text{LM}}$ with routers/LFF frozen. This step can further recover performance.

\section{Experiments}

\subsection{Experimental Setup}
Our experimental configuration aligns with SkipGPT's setting with the following specifications:

\textbf{Models.} 
We validate the proposed method on the open-source Llama~\citep{grattafiori2024llama} model with different scales, i.e. 3B and 8B.

\textbf{Data.}  The RedPajama-Data-1T-Sample~\citep{weber2024redpajama} corpus is utilized for both calibration and training.

\textbf{Evaluation Benchmarks.} 
Performance is assessed on:
\begin{itemize}
    \item \textit{Reasoning Tasks:} Accuracy on BoolQ \citep{clark2019boolq}, PIQA \citep{bisk2020piqa}, HellaSwag \citep{zellers2019hellaswag}, Winogrande \citep{sakaguchi2021winogrande}, ARC-E/C \citep{clark2018think}, and OBQA \citep{mihaylov-etal-2018-suit} via lm-evaluation-harness \citep{eval-harness}.
    \item \textit{Perplexity:} Perplexity (PPL) on WikiText-2 \citep{Merity2016PointerSM}.
\end{itemize}

\textbf{Baseline Methods.}
We selected state-of-the-art static pruning and dynamic computation allocation methods for comparison, with detailed descriptions in section \ref{sec:comparison_method}. Please refer to section ~\ref{sec:experiment_detail} in Appendix for detailed training/evaluating settings.


\subsection{Results}
\begin{table*}[t]
\centering
\caption{\small Performance comparison of different pruning methods on reasoning and language modeling tasks at sparsity levels of 25\% and 40\%. For reasoning tasks, we report accuracy (\%); higher is better. The average (AVG) accuracy across all reasoning tasks is included. For Wikitext-2 (WT2), we report perplexity (PPL); lower is better. The best results under each sparsity level are highlighted in \textbf{bold} and the second best are \underline{underlined}.}
\label{tab:main_results}

\begin{subtable}[h]{\textwidth}
\centering
\caption{Sparsity = 25\%}
\label{tab:results_25}
\tiny
\begin{tabular}{@{}l|cccccccc|c@{}}
\toprule
\multirow{2}{*}{Method} & \multicolumn{8}{c}{Reasoning (Acc. $\uparrow$)} & \multirow{2}{*}{WT2 (PPL $\downarrow$)} \\
\cmidrule{2-9}
 & BoolQ & OBQA & PIQA & WinoG. & Hella. & ARC-C & ARC-E & AVG & \\
\midrule
Dense & 82.14 & 44.6 & 81.07 & 77.43 & 81.89 & 57.68 & 84.81 & 72.80 & 7.33 \\
\midrule
\multicolumn{10}{l}{\textit{Static}} \\
SliceGPT & \textbf{72.39} & 34.4 & 66.7 & 61.56 & 56.96 & 31.48 & 50.08 & 53.37 & \underline{9.22} \\
Shortened-llama & 71.19 & 37.4 & 73.72 & \textbf{71.82} & 69.56 & 44.45 & 66.88 & 62.15 & 10.32 \\
ShortGPT & \underline{72.05} & 38.4 & 73.94 & \underline{70.96} & 69.23 & 43.86 & 68.01 & 62.35 & 11.13 \\
\midrule
\multicolumn{10}{l}{\textit{Dynamic}} \\
MoD & 50.28 & 31.6 & 64.25 & 52.41 & 50.44 & 28.24 & 37.67 & 44.98 & 34.21 \\
D-LLM & 50.36 & 30.2 & 57.4 & 52.49 & 37.64 & 28.16 & 37.12 & 41.91 & 40.12 \\
\rowcolor{lightgray}
SkipGPT-Router & 54.13 & 27.6 & 53.92 & 54.46 & 60.92 & 39.25 & 68.31 & 51.23 & 20.76 \\
\rowcolor{lightgray}
LFF-Router (ours) & 71.19 & \underline{40.80} & \underline{74.97} & 63.69 & 73.35& 49.23& \underline{79.08}& \underline{64.62} &9.43 \\
\rowcolor{lightyellow}
SkipGPT-Lora & 70.67 & 29.60 & 56.96 & 62.83 & \underline{74.22} & \underline{49.91} & 78.79 & 60.43 & 10.53 \\
\rowcolor{lightyellow}
LFF-Lora (ours) & 71.93 & \textbf{41.80}&\textbf{76.82} & 65.19 & \textbf{76.54} & \textbf{51.45} & \textbf{79.38} & \textbf{66.16}& \textbf{8.91} \\
\bottomrule
\end{tabular}
\end{subtable}

\vspace{1em}

\begin{subtable}[h]{\textwidth}
\centering
\caption{Sparsity = 40\%}
\label{tab:results_40}
\tiny
\begin{tabular}{@{}l|cccccccc|c@{}}
\toprule
\multirow{2}{*}{Method} & \multicolumn{8}{c}{Reasoning (Acc. $\uparrow$)} & \multirow{2}{*}{WT2 (PPL $\downarrow$)} \\
\cmidrule{2-9}
 & BoolQ & OBQA & PIQA & WinoG. & Hella. & ARC-C & ARC-E & AVG & \\
\midrule
Dense & 82.14 & 44.6 & 81.07 & 77.43 & 81.89 & 57.68 & 84.81 & 72.80 & 7.33 \\
\midrule
\multicolumn{10}{l}{\textit{Static}} \\
SliceGPT & \textbf{67.52} & 28.2 & 60.61 & 55.41 & 44.15 & 25.34 & 40.7 & 45.99 & 14.87 \\
Shortened-llama & 65.02 & 32.4 & 68.01 & \underline{64.64} & 57.55 & 33.02 & 53.11 & 53.39 & 17.22 \\
ShortGPT & 65.38 & 32.0 & 68.61 & \textbf{67.32} & 58.43 & 35.32 & 53.37 & 54.35 & 18.35 \\
\midrule
\multicolumn{10}{l}{\textit{Dynamic}} \\
MoD & 50.28 & 33.0 & 65.56 & 51.38 & 54.01 & 30.2 & 38.09 & 46.07 & 40.42 \\
D-LLM & 50.00 & 31.8 & 58.54 & 51.78 & 48.3 & 26.88 & 44.82 & 44.59 & 52.78 \\
\rowcolor{lightgray}
SkipGPT-Router & 53.82 & 31.8 & 60.23 & 54.22 & 46.02 & 28.75 & 52.44 & 46.75 & 53.94 \\
\rowcolor{lightgray}
LFF-Router (ours) & 64.43 & 36.0 & \underline{71.87} & 52.17 & 59.95 & 37.71 & 69.99 & 56.02 & \underline{13.87} \\
\rowcolor{lightyellow}
SkipGPT-LoRA & \underline{66.57} & \underline{37.6} & 70.78 & 56.75 & \underline{65.17} & \underline{42.66} & \underline{72.39} & \underline{58.85} & 14.35 \\
\rowcolor{lightyellow}
LFF-LoRA  (ours)& 65.99 & \textbf{38.0} & \textbf{73.39} & 58.8& \textbf{69.45} & \textbf{43.6}& \textbf{72.43} & \textbf{60.24} & \textbf{11.11} \\
\bottomrule
\end{tabular}
\end{subtable}
\end{table*}

We conduct extensive experiments to evaluate the proposed informed routing paradigm with LFF. Our results demonstrate its advantages in training stability, efficiency, and performance preservation compared to the traditional greedy routing approaches and static compressing methods. Notably, except for SkipGPT and LFF (ours), all methods report results from LoRA finetuned models. To further demonstrate the effectiveness of our method, we report two-phase results (router training + LoRA finetune) for SkipGPT and LFF, since SkipGPT essentially serves as an ablation experiment for the informed routing component in our method.

\paragraph{Inconsistency Between Language Modeling and Reasoning}

Experimental results reveal an inconsistency between compressed models’ language modeling (LM) capability and their reasoning performance. As shown in Table~\ref{tab:main_results}, while perplexity (PPL) trends generally align with reasoning accuracy, certain methods deviate. For instance, SliceGPT at 25\% sparsity ranks second in PPL but drops to sixth in average reasoning accuracy. Similarly, at 40\% sparsity, LFF-Router achieves better PPL than SkipGPT-LoRA yet shows a 3\% drop in reasoning. These indicate that LM loss alone may not fully reflect a compressed model’s reasoning ability, underscoring the need for evaluation across diverse task-specific benchmarks.

\paragraph{Training Stability and Efficiency Gains of LFF Initialization}

The proposed informed routing approach significantly improves training stability and efficiency by initializing the router with a pre-fit LFF. This leads to faster and smoother router convergence. Intuitively, after router training, at 25\% sparsity, LFF-Router reduces PPL by 11 points compared to SkipGPT-Router; at 40\% sparsity, the reduction reaches 40 points. Notably, LFF-Router at 25\% sparsity outperforms fully fine-tuned SkipGPT-LoRA while saving over 50\% training time (details can be find in section \ref{sec:experiment_detail}). 

\paragraph{Superiority After Fine-tuning and Underlying Mechanisms}

After LoRA fine-tuning, our method outperforms SkipGPT in 15 out of 16 tasks. We attribute this to two factors: \textbf{(1)} The LFF better preserves the original feature distribution by approximating the layer transformation instead of discarding tokens. Features processed by LFF show higher cosine similarity and lower L1 loss (0.16 vs. 0.52), providing a warmer start for fine-tuning. \textbf{(2)} Pre-fitting the LFF enables the router to prioritize tokens with high recoverability—those predictable by a simple network—leading to a healthier model structure and better parameter recovery during LoRA fine-tuning.

\subsection{Further Analysis}

\paragraph{Analysis of Router Behavior}
\label{subsec:router_behavior}

\begin{table*}[h]
\centering
\caption{\small Reduction ratio between Attention and MLP modules at different global sparsity levels.}
\label{tab:layer_sparsity}
\tiny
\begin{tabular}{lcccccc}
\toprule
\multirow{2}{*}{\textbf{Method}} & \multicolumn{2}{c}{\textbf{25\% Sparsity}} & \multicolumn{2}{c}{\textbf{40\% Sparsity}} & \multicolumn{2}{c}{\textbf{70\% Sparsity}} \\
\cmidrule(lr){2-3} \cmidrule(lr){4-5} \cmidrule(lr){6-7}
& Attention & MLP & Attention & MLP & Attention & MLP \\
\midrule
SkipGPT & 58.0\% & 42.0\% & 57.8\% & 42.2\% & 56.4\% & 43.6\% \\
LFF & 71.4\% & 28.6\% & 67.2\% & 32.8\% & 66.2\% & 33.8\% \\
\bottomrule
\end{tabular}
\end{table*}

As shown in Table~\ref{tab:layer_sparsity}, our method consistently select more tokens from self-attention modules than the SkipGPT baseline, across all sparsity levels. This supports prior findings \cite{he2024matterstransformersattentionneeded} that self-attention is more redundant than FFN blocks. The success of our lightweight, linear LFF in predicting attention outputs suggests that many token transformations in self-attention are approximable by simple linear operations. We term this property \textbf{linear simplicity}. Our router, preconditioned by the LFF, learns to identify such tokens, leading to a more explainable sparsity profile.

\paragraph{Balanced computation between attention and FFN}
A potential point of discussion is our treatment of self-attention and FFN modules as equally valid candidates for computation reduction, a design choice inherited from SkipGPT. While self-attention contains fewer parameters, its computational complexity scales quadratically with sequence length, often making it the dominant computational bottleneck in modern long-context large language models. To preemptively address any concern that a direct comparison might be unfair, we conducted a rigorous ablation study. In this experiment, we independently controlled and enforced identical sparsity levels for each module type—self-attention and FFN—across all layers. This ensures a perfectly equitable comparison of the routing strategies' efficiency on a per-module-class basis. The results, showed in Table~\ref{tab:balance},  demonstrate that our informed routing paradigm consistently achieves superior performance compared to the greedy routing baseline under these controlled sparsity conditions. This finding robustly confirms that the performance gain of our method is not an artifact of an imbalanced reduction strategy but is intrinsically linked to its preservation of features distributions, validating our core hypothesis.

\begin{table*}[htbp]
\centering
\caption{\small Performance of SkipGPT-balance and LFF-balance methods under balanced computation reduction setting. All results are reported via LoRA-finetuned model at 25\% sparsity.}
\label{tab:balance}
\tiny
\begin{tabular}{@{}l|cccccccc|c@{}}
\toprule
\multirow{2}{*}{Method} & \multicolumn{8}{c}{Reasoning (Acc. $\uparrow$)} & \multirow{2}{*}{WT2 (PPL $\downarrow$)} \\
\cmidrule{2-9}
 & BoolQ & OBQA & PIQA & WinoG. & Hella. & ARC-C & ARC-E & AVG & \\
\midrule
SkipGPT-balance & 67.77 & 39.60 & 62.08 & 60.93 & 71.33 & 40.44 & 66.84 & 58.42 & 11.40 \\
LFF-balance & \textbf{70.54} & \textbf{39.80} & \textbf{74.71} & \textbf{63.62}& \textbf{73.16} & \textbf{47.24}& \textbf{75.38} & \textbf{63.49} & \textbf{9.49} \\
\bottomrule
\end{tabular}
\end{table*}

\paragraph{Can informed routing increase the upper limit of computational reduction? }
\label{subsec:upper_limit}

\begin{table*}[h]
\centering
\caption{\small Performance of SkipGPT-LoRA and LFF-Router across different sparsity. Even without the final fine-tuning stage, LFF-Router can already surpass SkipGPT with both router and LLM fine-tuning.}
\label{tab:limit}
\tiny
\begin{tabular}{lcccccc}
\toprule
\multirow{2}{*}{\textbf{Method}} & \multicolumn{2}{c}{\textbf{25\% Sparsity}} & \multicolumn{2}{c}{\textbf{40\% Sparsity}} & \multicolumn{2}{c}{\textbf{70\% Sparsity}} \\
\cmidrule(lr){2-3} \cmidrule(lr){4-5} \cmidrule(lr){6-7}
& Val. Loss$\downarrow$ & PPL $\downarrow$ & Val. Loss$\downarrow$ & PPL$\downarrow$ & Val. Loss$\downarrow$ & PPL$\downarrow$ \\
\midrule
SkipGPT-LoRA & 2.38 & 9.61 & 2.74 & 14.34 & \textbf{3.72} & \textbf{52.75} \\
LFF-Router & \textbf{2.36} & \textbf{9.46} & \textbf{2.57} & \textbf{12.89} & 4.08 & 88.17 \\
\bottomrule
\end{tabular}
\end{table*}

We investigate whether the proposed \textit{informed routing} approach can elevate  the upper bound of computational reduction. Empirical results suggest otherwise. As shown in Table \ref{tab:limit}, at 25\% sparsity, LFF-Router \textit{(require only LFF initialization and router training, without LoRA finetuning)} surpasses the full training of SkipGPT-LoRA, indicating effective feature forecasting. However, at 40\% sparsity, LFF-Router fails to outperform  SkipGPT-LoRA in reasoning tasks, revealing its forecasting limits. At 70\% sparsity, the gap widens substantially in language modeling, confirming that LFF's capacity is exceeded. Thus, while effective at moderate sparsity, informed routing does not extend the ultimate computation reduction boundary.

\paragraph{Generalization on Llama-3B}
\begin{table}[htbp]
\centering
\caption{\small Performance comparison on Llama3.2-3B with 25\% sparsity. Accuracies (\%) on reasoning tasks; perplexity (PPL) on WikiText-2.}
\centering
\label{tab:llama3b}
\tiny
\begin{tabular}{@{}l|cccccccc|c@{}}
\toprule
\multirow{2}{*}{Method} & \multicolumn{8}{c}{Reasoning (Acc. $\uparrow$)} & \multirow{2}{*}{WT2 (PPL $\downarrow$)} \\
\cmidrule{2-9}
 & BoolQ & OBQA & PIQA & WinoG. & Hella. & ARC-C & ARC-E & AVG & \\
\midrule
Dense & 73.03 & 43.40 & 77.58 & 72.22 & 76.41 & 50.85 & 79.17 & 67.52 & 9.27 \\
\midrule
SkipGPT-Router & 47.65 & 34.40 & 61.15 & 54.14 & 51.99 & 26.62 & 39.56 & 45.07 & 36.50 \\
LFF-Router (ours) & 61.74 & 34.80 & \underline{68.28} & 58.33 & 62.36& \underline{40.53}& \underline{71.04}& 56.73 &\underline{12.19} \\
SkipGPT-LoRA & \underline{62.81} & \underline{36.60} & 66.49 & \textbf{59.19} & \underline{64.52} & 39.51 & 70.71 & \underline{57.12} & 14.82 \\
LFF-LoRA (ours) & \textbf{62.87} & \textbf{37.20}&\textbf{69.04} & \underline{59.04} & \textbf{66.14} & \textbf{41.81} & \textbf{73.06} & \textbf{58.45}& \textbf{11.46} \\
\bottomrule
\end{tabular}
\end{table}

To validate the generalization across model scales, we evaluate our method on the Llama3.2-3B model. As shown in Table~\ref{tab:llama3b}, the results align with those from the 8B model, substantiating our approach's efficacy. At 25\% sparsity, LFF-Router significantly outperforms SkipGPT-Router, improving the average accuracy on reasoning tasks by 11\% and reducing language modeling perplexity by 24. Moreover, LFF-Router matches the performance of SkipGPT-LoRA while saving over 50\% in training time. After LoRA fine-tuning, LFF-LoRA achieves superior performance on 8 out of 9 tasks, confirming the advantage of informed routing. An key finding is that the performance degradation is more pronounced on the 3B model, indicating its lower intrinsic redundancy and higher sensitivity to computation reduction.

\paragraph{Allocation Visualization for Attention and FFN Modules}\label{app:sparsity}

\begin{figure}[htbp]
\centering
\includegraphics[width=\linewidth]{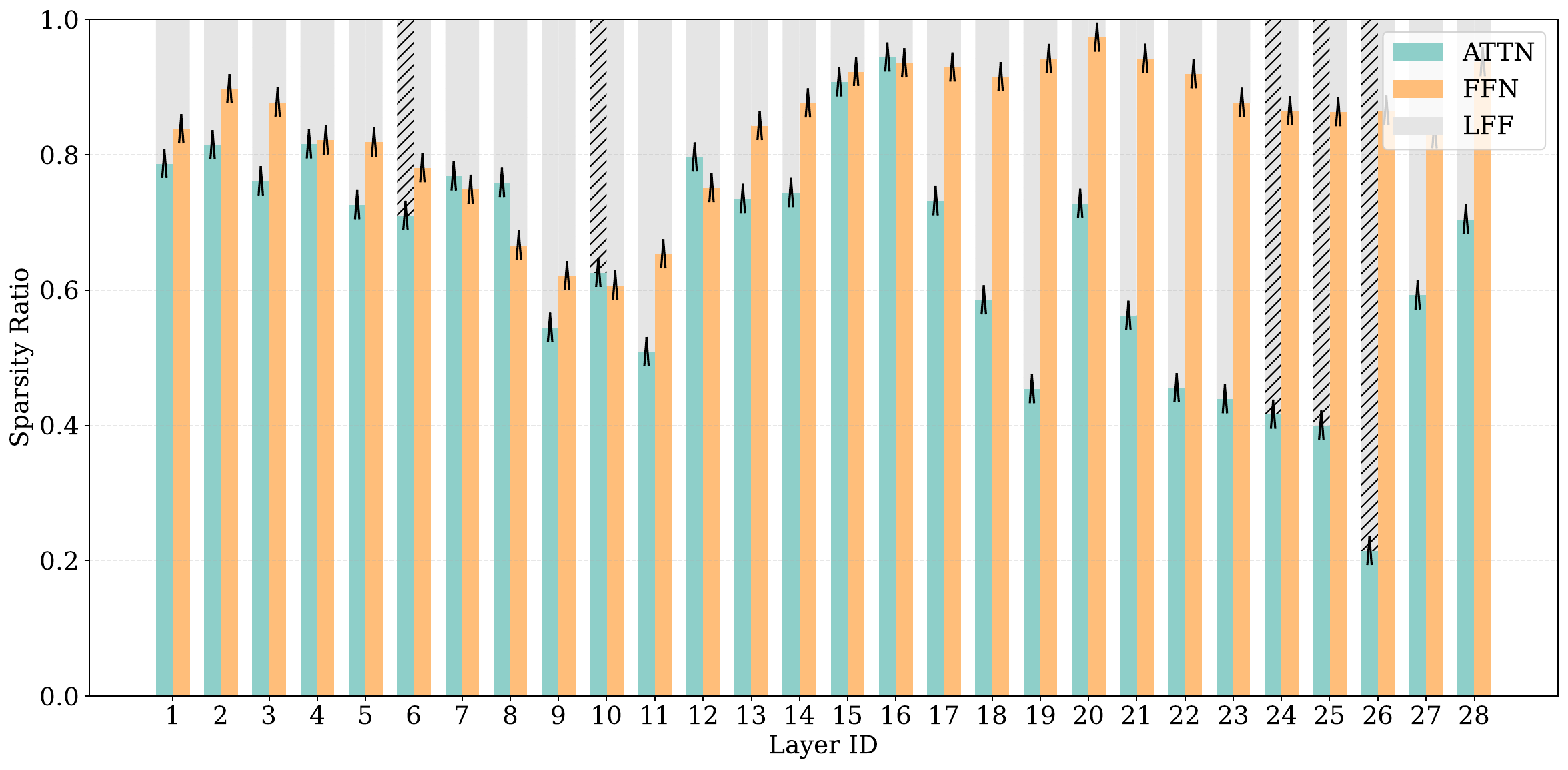}
\caption{\small Layer-wise Token Allocation. The hatched area represents the proportion of tokens processed by the efficient LFF branch and the colored areas show the tokens retained for full computation in the Attention (green) and FFN (orange) modules.}
\label{fig:layerwise_sparsity}
\end{figure}

This section presents an intuitive visualization of token allocation by routers across Attention (ATTN) and Feed-Forward Network (FFN) modules within the Llama-3B model. We track and record the layer-wise allocation information for each batch on the validation set, and then present the average value across all batches.

As shown in Figure~\ref{fig:layerwise_sparsity}, which details the allocation across 28 layers, the routing mechanism intelligently distributes input tokens, creating a dynamic computational sparsity pattern. A key observation is that the proportion of tokens directed to the LFF branch (hatched area) varies significantly across layers (also between attention and FFN modules), suggesting that the router adapts its filtering strategy based on the hierarchical processing needs of the network. This strategic allocation preserves critical tokens for the full computational pipeline while efficiently processing others, effectively reducing the overall computational overhead without compromising performance.

\section{Conclusion}
In this work, we identified fundamental limitations in the established greedy routing paradigm for dynamic computation reduction in large language models: its reactive nature leads to irreversible information loss and its token selection criterion is inherently short-sighted. In response, we proposed a paradigm-shifting alternative, informed routing, which introduces Lightweight Feature Forecasters to fit inter-layer transformations before routing decisions are made.
Our approach offers key advantages: First, LFFs approximate skipped tokens, reducing feature shift and improving initial stability with less performance drop. Second, LFF forecasting error gives the router a recoverability-based importance measure, enabling smarter retention of hard-to-predict tokens. Third, our method consistently outperform greedy routing methods on both unbalanced and balanced reduction setting. Finally, we show self-attention's redundancy stems from linearly approximable transformations.
Despite these advancements, our exploration of extreme sparsity levels (e.g., 70\%) reveals that the upper limit of dynamic computation allocation is ultimately governed by the complexity of the underlying transformations, which a simple LFF cannot fully capture. This presents an exciting avenue for future work, which could explore more sophisticated yet efficient forecasters or hybrid strategies.

\bibliography{iclr2026_conference}

\begin{thebibliography}{37}
\providecommand{\natexlab}[1]{#1}
\providecommand{\url}[1]{\texttt{#1}}
\expandafter\ifx\csname urlstyle\endcsname\relax
  \providecommand{\doi}[1]{doi: #1}\else
  \providecommand{\doi}{doi: \begingroup \urlstyle{rm}\Url}\fi

\bibitem[Ashkboos et~al.(2024)Ashkboos, Croci, do~Nascimento, Hoefler, and
  Hensman]{ashkboos2024slicegpt}
Saleh Ashkboos, Maximilian~L. Croci, Marcelo~Gennari do~Nascimento, Torsten
  Hoefler, and James Hensman.
\newblock Slice{GPT}: Compress large language models by deleting rows and
  columns.
\newblock In \emph{The Twelfth International Conference on Learning
  Representations}, 2024.
\newblock URL \url{https://openreview.net/forum?id=vXxardq6db}.

\bibitem[Bisk et~al.(2020)Bisk, Zellers, Gao, Choi, et~al.]{bisk2020piqa}
Yonatan Bisk, Rowan Zellers, Jianfeng Gao, Yejin Choi, et~al.
\newblock Piqa: Reasoning about physical commonsense in natural language.
\newblock In \emph{Proceedings of the AAAI conference on artificial
  intelligence}, volume~34, pp.\  7432--7439, 2020.

\bibitem[Cai et~al.(2025)Cai, Li, Wang, Zhu, Shen, Li, and Chua]{cai2025large}
Hongru Cai, Yongqi Li, Wenjie Wang, Fengbin Zhu, Xiaoyu Shen, Wenjie Li, and
  Tat-Seng Chua.
\newblock Large language models empowered personalized web agents.
\newblock In \emph{Proceedings of the ACM on Web Conference 2025}, pp.\
  198--215, 2025.

\bibitem[Chen et~al.(2025{\natexlab{a}})Chen, Hu, Zhang, Wang, Li, and
  Chen]{ICLR2025_4b00a351}
Xiaodong Chen, Yuxuan Hu, Jing Zhang, Yanling Wang, Cuiping Li, and Hong Chen.
\newblock Streamlining redundant layers to compress large language models.
\newblock In Y.~Yue, A.~Garg, N.~Peng, F.~Sha, and R.~Yu (eds.),
  \emph{International Conference on Representation Learning}, volume 2025, pp.\
   30362--30383, 2025{\natexlab{a}}.
\newblock URL
  \url{https://proceedings.iclr.cc/paper_files/paper/2025/file/4b00a351b41358965613c118e87dc28c-Paper-Conference.pdf}.

\bibitem[Chen et~al.(2025{\natexlab{b}})Chen, Zhang, Zeng, Wei, Wang, Ling, Li,
  and Yuan]{chen2025prune}
Xinrui Chen, Hongxing Zhang, Fanyi Zeng, Yongxian Wei, Yizhi Wang, Xitong Ling,
  Guanghao Li, and Chun Yuan.
\newblock Prune\&comp: Free lunch for layer-pruned llms via iterative pruning
  with magnitude compensation.
\newblock \emph{arXiv preprint arXiv:2507.18212}, 2025{\natexlab{b}}.

\bibitem[Clark et~al.(2019)Clark, Lee, Chang, Kwiatkowski, Collins, and
  Toutanova]{clark2019boolq}
Christopher Clark, Kenton Lee, Ming-Wei Chang, Tom Kwiatkowski, Michael
  Collins, and Kristina Toutanova.
\newblock Boolq: Exploring the surprising difficulty of natural yes/no
  questions.
\newblock \emph{arXiv preprint arXiv:1905.10044}, 2019.

\bibitem[Clark et~al.(2018)Clark, Cowhey, Etzioni, Khot, Sabharwal, Schoenick,
  and Tafjord]{clark2018think}
Peter Clark, Isaac Cowhey, Oren Etzioni, Tushar Khot, Ashish Sabharwal, Carissa
  Schoenick, and Oyvind Tafjord.
\newblock Think you have solved question answering? try arc, the ai2 reasoning
  challenge.
\newblock \emph{arXiv preprint arXiv:1803.05457}, 2018.

\bibitem[DeepSeek-AI et~al.(2025)DeepSeek-AI, Guo, Yang, Zhang, Song, Zhang,
  Xu, Zhu, Ma, Wang, Bi, Zhang, Yu, Wu, Wu, Gou, Shao, Li, Gao, Liu, Xue, Wang,
  Wu, Feng, Lu, Zhao, Deng, Zhang, Ruan, Dai, Chen, Ji, Li, Lin, Dai, Luo, Hao,
  Chen, Li, Zhang, Bao, Xu, Wang, Ding, Xin, Gao, Qu, Li, Guo, Li, Wang, Chen,
  Yuan, Qiu, Li, Cai, Ni, Liang, Chen, Dong, Hu, Gao, Guan, Huang, Yu, Wang,
  Zhang, Zhao, Wang, Zhang, Xu, Xia, Zhang, Zhang, Tang, Li, Wang, Li, Tian,
  Huang, Zhang, Wang, Chen, Du, Ge, Zhang, Pan, Wang, Chen, Jin, Chen, Lu,
  Zhou, Chen, Ye, Wang, Yu, Zhou, Pan, Li, Zhou, Wu, Ye, Yun, Pei, Sun, Wang,
  Zeng, Zhao, Liu, Liang, Gao, Yu, Zhang, Xiao, An, Liu, Wang, Chen, Nie,
  Cheng, Liu, Xie, Liu, Yang, Li, Su, Lin, Li, Jin, Shen, Chen, Sun, Wang,
  Song, Zhou, Wang, Shan, Li, Wang, Wei, Zhang, Xu, Li, Zhao, Sun, Wang, Yu,
  Zhang, Shi, Xiong, He, Piao, Wang, Tan, Ma, Liu, Guo, Ou, Wang, Gong, Zou,
  He, Xiong, Luo, You, Liu, Zhou, Zhu, Xu, Huang, Li, Zheng, Zhu, Ma, Tang,
  Zha, Yan, Ren, Ren, Sha, Fu, Xu, Xie, Zhang, Hao, Ma, Yan, Wu, Gu, Zhu, Liu,
  Li, Xie, Song, Pan, Huang, Xu, Zhang, and
  Zhang]{deepseekai2025deepseekr1incentivizingreasoningcapability}
DeepSeek-AI, Daya Guo, Dejian Yang, Haowei Zhang, Junxiao Song, Ruoyu Zhang,
  Runxin Xu, Qihao Zhu, Shirong Ma, Peiyi Wang, Xiao Bi, Xiaokang Zhang,
  Xingkai Yu, Yu~Wu, Z.~F. Wu, Zhibin Gou, Zhihong Shao, Zhuoshu Li, Ziyi Gao,
  Aixin Liu, Bing Xue, Bingxuan Wang, Bochao Wu, Bei Feng, Chengda Lu,
  Chenggang Zhao, Chengqi Deng, Chenyu Zhang, Chong Ruan, Damai Dai, Deli Chen,
  Dongjie Ji, Erhang Li, Fangyun Lin, Fucong Dai, Fuli Luo, Guangbo Hao,
  Guanting Chen, Guowei Li, H.~Zhang, Han Bao, Hanwei Xu, Haocheng Wang,
  Honghui Ding, Huajian Xin, Huazuo Gao, Hui Qu, Hui Li, Jianzhong Guo, Jiashi
  Li, Jiawei Wang, Jingchang Chen, Jingyang Yuan, Junjie Qiu, Junlong Li, J.~L.
  Cai, Jiaqi Ni, Jian Liang, Jin Chen, Kai Dong, Kai Hu, Kaige Gao, Kang Guan,
  Kexin Huang, Kuai Yu, Lean Wang, Lecong Zhang, Liang Zhao, Litong Wang, Liyue
  Zhang, Lei Xu, Leyi Xia, Mingchuan Zhang, Minghua Zhang, Minghui Tang, Meng
  Li, Miaojun Wang, Mingming Li, Ning Tian, Panpan Huang, Peng Zhang, Qiancheng
  Wang, Qinyu Chen, Qiushi Du, Ruiqi Ge, Ruisong Zhang, Ruizhe Pan, Runji Wang,
  R.~J. Chen, R.~L. Jin, Ruyi Chen, Shanghao Lu, Shangyan Zhou, Shanhuang Chen,
  Shengfeng Ye, Shiyu Wang, Shuiping Yu, Shunfeng Zhou, Shuting Pan, S.~S. Li,
  Shuang Zhou, Shaoqing Wu, Shengfeng Ye, Tao Yun, Tian Pei, Tianyu Sun,
  T.~Wang, Wangding Zeng, Wanjia Zhao, Wen Liu, Wenfeng Liang, Wenjun Gao,
  Wenqin Yu, Wentao Zhang, W.~L. Xiao, Wei An, Xiaodong Liu, Xiaohan Wang,
  Xiaokang Chen, Xiaotao Nie, Xin Cheng, Xin Liu, Xin Xie, Xingchao Liu, Xinyu
  Yang, Xinyuan Li, Xuecheng Su, Xuheng Lin, X.~Q. Li, Xiangyue Jin, Xiaojin
  Shen, Xiaosha Chen, Xiaowen Sun, Xiaoxiang Wang, Xinnan Song, Xinyi Zhou,
  Xianzu Wang, Xinxia Shan, Y.~K. Li, Y.~Q. Wang, Y.~X. Wei, Yang Zhang,
  Yanhong Xu, Yao Li, Yao Zhao, Yaofeng Sun, Yaohui Wang, Yi~Yu, Yichao Zhang,
  Yifan Shi, Yiliang Xiong, Ying He, Yishi Piao, Yisong Wang, Yixuan Tan,
  Yiyang Ma, Yiyuan Liu, Yongqiang Guo, Yuan Ou, Yuduan Wang, Yue Gong, Yuheng
  Zou, Yujia He, Yunfan Xiong, Yuxiang Luo, Yuxiang You, Yuxuan Liu, Yuyang
  Zhou, Y.~X. Zhu, Yanhong Xu, Yanping Huang, Yaohui Li, Yi~Zheng, Yuchen Zhu,
  Yunxian Ma, Ying Tang, Yukun Zha, Yuting Yan, Z.~Z. Ren, Zehui Ren, Zhangli
  Sha, Zhe Fu, Zhean Xu, Zhenda Xie, Zhengyan Zhang, Zhewen Hao, Zhicheng Ma,
  Zhigang Yan, Zhiyu Wu, Zihui Gu, Zijia Zhu, Zijun Liu, Zilin Li, Ziwei Xie,
  Ziyang Song, Zizheng Pan, Zhen Huang, Zhipeng Xu, Zhongyu Zhang, and Zhen
  Zhang.
\newblock Deepseek-r1: Incentivizing reasoning capability in llms via
  reinforcement learning, 2025.
\newblock URL \url{https://arxiv.org/abs/2501.12948}.

\bibitem[Gao et~al.(2024)Gao, Tow, Abbasi, Biderman, Black, DiPofi, Foster,
  Golding, Hsu, Le~Noac'h, Li, McDonell, Muennighoff, Ociepa, Phang, Reynolds,
  Schoelkopf, Skowron, Sutawika, Tang, Thite, Wang, Wang, and
  Zou]{eval-harness}
Leo Gao, Jonathan Tow, Baber Abbasi, Stella Biderman, Sid Black, Anthony
  DiPofi, Charles Foster, Laurence Golding, Jeffrey Hsu, Alain Le~Noac'h,
  Haonan Li, Kyle McDonell, Niklas Muennighoff, Chris Ociepa, Jason Phang,
  Laria Reynolds, Hailey Schoelkopf, Aviya Skowron, Lintang Sutawika, Eric
  Tang, Anish Thite, Ben Wang, Kevin Wang, and Andy Zou.
\newblock The language model evaluation harness, 07 2024.
\newblock URL \url{https://zenodo.org/records/12608602}.

\bibitem[Grattafiori et~al.(2024)Grattafiori, Dubey, Jauhri, Pandey, Kadian,
  Al-Dahle, Letman, Mathur, Schelten, Vaughan, et~al.]{grattafiori2024llama}
Aaron Grattafiori, Abhimanyu Dubey, Abhinav Jauhri, Abhinav Pandey, Abhishek
  Kadian, Ahmad Al-Dahle, Aiesha Letman, Akhil Mathur, Alan Schelten, Alex
  Vaughan, et~al.
\newblock The llama 3 herd of models.
\newblock \emph{arXiv preprint arXiv:2407.21783}, 2024.

\bibitem[Han et~al.(2016)Han, Mao, and Dally]{han2016deep}
Song Han, Huizi Mao, and William~J. Dally.
\newblock Deep compression: Compressing deep neural network with pruning,
  trained quantization and huffman coding.
\newblock In \emph{ICLR}, 2016.
\newblock URL \url{http://arxiv.org/abs/1510.00149}.

\bibitem[He \& Lin(2025)He and Lin]{he2025olica}
Jiujun He and Huazhen Lin.
\newblock Olica: Efficient structured pruning of large language models without
  retraining.
\newblock \emph{arXiv preprint arXiv:2506.08436}, 2025.

\bibitem[He et~al.(2024)He, Sun, Shen, and
  Li]{he2024matterstransformersattentionneeded}
Shwai He, Guoheng Sun, Zheyu Shen, and Ang Li.
\newblock What matters in transformers? not all attention is needed, 2024.
\newblock URL \url{https://arxiv.org/abs/2406.15786}.

\bibitem[Hu et~al.(2022)Hu, Shen, Wallis, Allen-Zhu, Li, Wang, Wang, Chen,
  et~al.]{hu2022lora}
Edward~J Hu, Yelong Shen, Phillip Wallis, Zeyuan Allen-Zhu, Yuanzhi Li, Shean
  Wang, Lu~Wang, Weizhu Chen, et~al.
\newblock Lora: Low-rank adaptation of large language models.
\newblock \emph{ICLR}, 1\penalty0 (2):\penalty0 3, 2022.

\bibitem[Ji et~al.(2025)Ji, Zhang, Xia, Chen, Shou, Chen, and
  Li]{ji2025SpecVLM}
Yicheng Ji, Jun Zhang, Heming Xia, Jinpeng Chen, Lidan Shou, Gang Chen, and
  Huan Li.
\newblock Specvlm: Enhancing speculative decoding of video llms via
  verifier-guided token pruning.
\newblock In \emph{The 2025 Conference on Empirical Methods in Natural Language
  Processing}, 2025.
\newblock URL \url{https://openreview.net/forum?id=mWElG6fKEN}.

\bibitem[Jiang et~al.(2024)Jiang, Wang, Xie, Zhao, Qian, Lui,
  et~al.]{jiang2024d}
Yikun Jiang, Huanyu Wang, Lei Xie, Hanbin Zhao, Hui Qian, John Lui, et~al.
\newblock D-llm: A token adaptive computing resource allocation strategy for
  large language models.
\newblock \emph{Advances in Neural Information Processing Systems},
  37:\penalty0 1725--1749, 2024.

\bibitem[Kaplan et~al.(2020)Kaplan, McCandlish, Henighan, Brown, Chess, Child,
  Gray, Radford, Wu, and Amodei]{kaplan2020scalinglawsneurallanguage}
Jared Kaplan, Sam McCandlish, Tom Henighan, Tom~B. Brown, Benjamin Chess, Rewon
  Child, Scott Gray, Alec Radford, Jeffrey Wu, and Dario Amodei.
\newblock Scaling laws for neural language models, 2020.
\newblock URL \url{https://arxiv.org/abs/2001.08361}.

\bibitem[Kim et~al.(2024)Kim, Kim, Kim, Castells, Choi, Shin, and
  Song]{kim2024shortened}
Bo-Kyeong Kim, Geonmin Kim, Tae-Ho Kim, Thibault Castells, Shinkook Choi, Junho
  Shin, and Hyoung-Kyu Song.
\newblock Shortened {LL}a{MA}: A simple depth pruning for large language
  models.
\newblock In \emph{ICLR 2024 Workshop on Mathematical and Empirical
  Understanding of Foundation Models}, 2024.
\newblock URL \url{https://openreview.net/forum?id=18VGxuOdpu}.

\bibitem[Lee et~al.(2025)Lee, Ramachandran, and Krishna]{lee2025recap}
Mingyu Lee, Akshat Ramachandran, and Tushar Krishna.
\newblock Recap: Training-free compensation for coarse activation channel
  pruning in compressed llms.
\newblock In \emph{Machine Learning for Computer Architecture and Systems},
  2025.

\bibitem[Loshchilov \& Hutter(2017)Loshchilov and
  Hutter]{loshchilov2017decoupled}
Ilya Loshchilov and Frank Hutter.
\newblock Decoupled weight decay regularization.
\newblock \emph{arXiv preprint arXiv:1711.05101}, 2017.

\bibitem[Ma et~al.(2023)Ma, Fang, and Wang]{ma2023llm}
Xinyin Ma, Gongfan Fang, and Xinchao Wang.
\newblock Llm-pruner: On the structural pruning of large language models.
\newblock \emph{Advances in neural information processing systems},
  36:\penalty0 21702--21720, 2023.

\bibitem[Men et~al.(2024)Men, Xu, Zhang, Wang, Lin, Lu, Han, and
  Chen]{men2024shortgptlayerslargelanguage}
Xin Men, Mingyu Xu, Qingyu Zhang, Bingning Wang, Hongyu Lin, Yaojie Lu, Xianpei
  Han, and Weipeng Chen.
\newblock Shortgpt: Layers in large language models are more redundant than you
  expect, 2024.
\newblock URL \url{https://arxiv.org/abs/2403.03853}.

\bibitem[Merity et~al.(2017)Merity, Xiong, Bradbury, and
  Socher]{Merity2016PointerSM}
Stephen Merity, Caiming Xiong, James Bradbury, and Richard Socher.
\newblock Pointer sentinel mixture models.
\newblock In \emph{ICLR}, volume abs/1609.07843, 2017.
\newblock URL \url{https://api.semanticscholar.org/CorpusID:16299141}.

\bibitem[Mihaylov et~al.(2018)Mihaylov, Clark, Khot, and
  Sabharwal]{mihaylov-etal-2018-suit}
Todor Mihaylov, Peter Clark, Tushar Khot, and Ashish Sabharwal.
\newblock Can a suit of armor conduct electricity? a new dataset for open book
  question answering.
\newblock In Ellen Riloff, David Chiang, Julia Hockenmaier, and Jun{'}ichi
  Tsujii (eds.), \emph{Proceedings of the 2018 Conference on Empirical Methods
  in Natural Language Processing}, pp.\  2381--2391, Brussels, Belgium,
  October-November 2018. Association for Computational Linguistics.
\newblock \doi{10.18653/v1/D18-1260}.
\newblock URL \url{https://aclanthology.org/D18-1260/}.

\bibitem[OpenAI et~al.(2024)OpenAI, Achiam, Adler, Agarwal, Ahmad, Akkaya,
  Aleman, Almeida, Altenschmidt, Altman, Anadkat, Avila, Babuschkin, Balaji,
  Balcom, Baltescu, Bao, Bavarian, Belgum, Bello, Berdine, Bernadett-Shapiro,
  Berner, Bogdonoff, Boiko, Boyd, Brakman, Brockman, Brooks, Brundage, Button,
  Cai, Campbell, Cann, Carey, Carlson, Carmichael, Chan, Chang, Chantzis, Chen,
  Chen, Chen, Chen, Chen, Chess, Cho, Chu, Chung, Cummings, Currier, Dai,
  Decareaux, Degry, Deutsch, Deville, Dhar, Dohan, Dowling, Dunning, Ecoffet,
  Eleti, Eloundou, Farhi, Fedus, Felix, Fishman, Forte, Fulford, Gao, Georges,
  Gibson, Goel, Gogineni, Goh, Gontijo-Lopes, Gordon, Grafstein, Gray, Greene,
  Gross, Gu, Guo, Hallacy, Han, Harris, He, Heaton, Heidecke, Hesse, Hickey,
  Hickey, Hoeschele, Houghton, Hsu, Hu, Hu, Huizinga, Jain, Jain, Jang, Jiang,
  Jiang, Jin, Jin, Jomoto, Jonn, Jun, Kaftan, Łukasz Kaiser, Kamali,
  Kanitscheider, Keskar, Khan, Kilpatrick, Kim, Kim, Kim, Kirchner, Kiros,
  Knight, Kokotajlo, Łukasz Kondraciuk, Kondrich, Konstantinidis, Kosic,
  Krueger, Kuo, Lampe, Lan, Lee, Leike, Leung, Levy, Li, Lim, Lin, Lin, Litwin,
  Lopez, Lowe, Lue, Makanju, Malfacini, Manning, Markov, Markovski, Martin,
  Mayer, Mayne, McGrew, McKinney, McLeavey, McMillan, McNeil, Medina, Mehta,
  Menick, Metz, Mishchenko, Mishkin, Monaco, Morikawa, Mossing, Mu, Murati,
  Murk, Mély, Nair, Nakano, Nayak, Neelakantan, Ngo, Noh, Ouyang, O'Keefe,
  Pachocki, Paino, Palermo, Pantuliano, Parascandolo, Parish, Parparita,
  Passos, Pavlov, Peng, Perelman, de~Avila Belbute~Peres, Petrov,
  de~Oliveira~Pinto, Michael, Pokorny, Pokrass, Pong, Powell, Power, Power,
  Proehl, Puri, Radford, Rae, Ramesh, Raymond, Real, Rimbach, Ross, Rotsted,
  Roussez, Ryder, Saltarelli, Sanders, Santurkar, Sastry, Schmidt, Schnurr,
  Schulman, Selsam, Sheppard, Sherbakov, Shieh, Shoker, Shyam, Sidor, Sigler,
  Simens, Sitkin, Slama, Sohl, Sokolowsky, Song, Staudacher, Such, Summers,
  Sutskever, Tang, Tezak, Thompson, Tillet, Tootoonchian, Tseng, Tuggle,
  Turley, Tworek, Uribe, Vallone, Vijayvergiya, Voss, Wainwright, Wang, Wang,
  Wang, Ward, Wei, Weinmann, Welihinda, Welinder, Weng, Weng, Wiethoff,
  Willner, Winter, Wolrich, Wong, Workman, Wu, Wu, Wu, Xiao, Xu, Yoo, Yu, Yuan,
  Zaremba, Zellers, Zhang, Zhang, Zhao, Zheng, Zhuang, Zhuk, and
  Zoph]{openai2024gpt4technicalreport}
OpenAI, Josh Achiam, Steven Adler, Sandhini Agarwal, Lama Ahmad, Ilge Akkaya,
  Florencia~Leoni Aleman, Diogo Almeida, Janko Altenschmidt, Sam Altman,
  Shyamal Anadkat, Red Avila, Igor Babuschkin, Suchir Balaji, Valerie Balcom,
  Paul Baltescu, Haiming Bao, Mohammad Bavarian, Jeff Belgum, Irwan Bello, Jake
  Berdine, Gabriel Bernadett-Shapiro, Christopher Berner, Lenny Bogdonoff, Oleg
  Boiko, Madelaine Boyd, Anna-Luisa Brakman, Greg Brockman, Tim Brooks, Miles
  Brundage, Kevin Button, Trevor Cai, Rosie Campbell, Andrew Cann, Brittany
  Carey, Chelsea Carlson, Rory Carmichael, Brooke Chan, Che Chang, Fotis
  Chantzis, Derek Chen, Sully Chen, Ruby Chen, Jason Chen, Mark Chen, Ben
  Chess, Chester Cho, Casey Chu, Hyung~Won Chung, Dave Cummings, Jeremiah
  Currier, Yunxing Dai, Cory Decareaux, Thomas Degry, Noah Deutsch, Damien
  Deville, Arka Dhar, David Dohan, Steve Dowling, Sheila Dunning, Adrien
  Ecoffet, Atty Eleti, Tyna Eloundou, David Farhi, Liam Fedus, Niko Felix,
  Simón~Posada Fishman, Juston Forte, Isabella Fulford, Leo Gao, Elie Georges,
  Christian Gibson, Vik Goel, Tarun Gogineni, Gabriel Goh, Rapha Gontijo-Lopes,
  Jonathan Gordon, Morgan Grafstein, Scott Gray, Ryan Greene, Joshua Gross,
  Shixiang~Shane Gu, Yufei Guo, Chris Hallacy, Jesse Han, Jeff Harris, Yuchen
  He, Mike Heaton, Johannes Heidecke, Chris Hesse, Alan Hickey, Wade Hickey,
  Peter Hoeschele, Brandon Houghton, Kenny Hsu, Shengli Hu, Xin Hu, Joost
  Huizinga, Shantanu Jain, Shawn Jain, Joanne Jang, Angela Jiang, Roger Jiang,
  Haozhun Jin, Denny Jin, Shino Jomoto, Billie Jonn, Heewoo Jun, Tomer Kaftan,
  Łukasz Kaiser, Ali Kamali, Ingmar Kanitscheider, Nitish~Shirish Keskar,
  Tabarak Khan, Logan Kilpatrick, Jong~Wook Kim, Christina Kim, Yongjik Kim,
  Jan~Hendrik Kirchner, Jamie Kiros, Matt Knight, Daniel Kokotajlo, Łukasz
  Kondraciuk, Andrew Kondrich, Aris Konstantinidis, Kyle Kosic, Gretchen
  Krueger, Vishal Kuo, Michael Lampe, Ikai Lan, Teddy Lee, Jan Leike, Jade
  Leung, Daniel Levy, Chak~Ming Li, Rachel Lim, Molly Lin, Stephanie Lin,
  Mateusz Litwin, Theresa Lopez, Ryan Lowe, Patricia Lue, Anna Makanju, Kim
  Malfacini, Sam Manning, Todor Markov, Yaniv Markovski, Bianca Martin, Katie
  Mayer, Andrew Mayne, Bob McGrew, Scott~Mayer McKinney, Christine McLeavey,
  Paul McMillan, Jake McNeil, David Medina, Aalok Mehta, Jacob Menick, Luke
  Metz, Andrey Mishchenko, Pamela Mishkin, Vinnie Monaco, Evan Morikawa, Daniel
  Mossing, Tong Mu, Mira Murati, Oleg Murk, David Mély, Ashvin Nair, Reiichiro
  Nakano, Rajeev Nayak, Arvind Neelakantan, Richard Ngo, Hyeonwoo Noh, Long
  Ouyang, Cullen O'Keefe, Jakub Pachocki, Alex Paino, Joe Palermo, Ashley
  Pantuliano, Giambattista Parascandolo, Joel Parish, Emy Parparita, Alex
  Passos, Mikhail Pavlov, Andrew Peng, Adam Perelman, Filipe de~Avila
  Belbute~Peres, Michael Petrov, Henrique~Ponde de~Oliveira~Pinto, Michael,
  Pokorny, Michelle Pokrass, Vitchyr~H. Pong, Tolly Powell, Alethea Power,
  Boris Power, Elizabeth Proehl, Raul Puri, Alec Radford, Jack Rae, Aditya
  Ramesh, Cameron Raymond, Francis Real, Kendra Rimbach, Carl Ross, Bob
  Rotsted, Henri Roussez, Nick Ryder, Mario Saltarelli, Ted Sanders, Shibani
  Santurkar, Girish Sastry, Heather Schmidt, David Schnurr, John Schulman,
  Daniel Selsam, Kyla Sheppard, Toki Sherbakov, Jessica Shieh, Sarah Shoker,
  Pranav Shyam, Szymon Sidor, Eric Sigler, Maddie Simens, Jordan Sitkin,
  Katarina Slama, Ian Sohl, Benjamin Sokolowsky, Yang Song, Natalie Staudacher,
  Felipe~Petroski Such, Natalie Summers, Ilya Sutskever, Jie Tang, Nikolas
  Tezak, Madeleine~B. Thompson, Phil Tillet, Amin Tootoonchian, Elizabeth
  Tseng, Preston Tuggle, Nick Turley, Jerry Tworek, Juan Felipe~Cerón Uribe,
  Andrea Vallone, Arun Vijayvergiya, Chelsea Voss, Carroll Wainwright,
  Justin~Jay Wang, Alvin Wang, Ben Wang, Jonathan Ward, Jason Wei, CJ~Weinmann,
  Akila Welihinda, Peter Welinder, Jiayi Weng, Lilian Weng, Matt Wiethoff, Dave
  Willner, Clemens Winter, Samuel Wolrich, Hannah Wong, Lauren Workman, Sherwin
  Wu, Jeff Wu, Michael Wu, Kai Xiao, Tao Xu, Sarah Yoo, Kevin Yu, Qiming Yuan,
  Wojciech Zaremba, Rowan Zellers, Chong Zhang, Marvin Zhang, Shengjia Zhao,
  Tianhao Zheng, Juntang Zhuang, William Zhuk, and Barret Zoph.
\newblock Gpt-4 technical report, 2024.
\newblock URL \url{https://arxiv.org/abs/2303.08774}.

\bibitem[Raposo et~al.(2024)Raposo, Ritter, Richards, Lillicrap, Humphreys, and
  Santoro]{raposo2024mixture}
David Raposo, Sam Ritter, Blake Richards, Timothy Lillicrap, Peter~Conway
  Humphreys, and Adam Santoro.
\newblock Mixture-of-depths: Dynamically allocating compute in
  transformer-based language models.
\newblock \emph{arXiv preprint arXiv:2404.02258}, 2024.

\bibitem[Rozière et~al.(2024)Rozière, Gehring, Gloeckle, Sootla, Gat, Tan,
  Adi, Liu, Sauvestre, Remez, Rapin, Kozhevnikov, Evtimov, Bitton, Bhatt,
  Ferrer, Grattafiori, Xiong, Défossez, Copet, Azhar, Touvron, Martin,
  Usunier, Scialom, and Synnaeve]{rozière2024codellamaopenfoundation}
Baptiste Rozière, Jonas Gehring, Fabian Gloeckle, Sten Sootla, Itai Gat,
  Xiaoqing~Ellen Tan, Yossi Adi, Jingyu Liu, Romain Sauvestre, Tal Remez,
  Jérémy Rapin, Artyom Kozhevnikov, Ivan Evtimov, Joanna Bitton, Manish
  Bhatt, Cristian~Canton Ferrer, Aaron Grattafiori, Wenhan Xiong, Alexandre
  Défossez, Jade Copet, Faisal Azhar, Hugo Touvron, Louis Martin, Nicolas
  Usunier, Thomas Scialom, and Gabriel Synnaeve.
\newblock Code llama: Open foundation models for code, 2024.
\newblock URL \url{https://arxiv.org/abs/2308.12950}.

\bibitem[Sakaguchi et~al.(2021)Sakaguchi, Bras, Bhagavatula, and
  Choi]{sakaguchi2021winogrande}
Keisuke Sakaguchi, Ronan~Le Bras, Chandra Bhagavatula, and Yejin Choi.
\newblock Winogrande: An adversarial winograd schema challenge at scale.
\newblock \emph{Communications of the ACM}, 64\penalty0 (9):\penalty0 99--106,
  2021.

\bibitem[Shin et~al.(2025)Shin, Oh, and Oh]{shin2025orthorank}
Seungjun Shin, Jaehoon Oh, and Dokwan Oh.
\newblock Orthorank: Token selection via sink token orthogonality for efficient
  llm inference.
\newblock \emph{arXiv preprint arXiv:2507.03865}, 2025.

\bibitem[Su et~al.(2022)Su, Zhou, Yu, Shen, Chen, Zhu, Yu, and
  Zhou]{su2022welm}
Hui Su, Xiao Zhou, Houjin Yu, Xiaoyu Shen, Yuwen Chen, Zilin Zhu, Yang Yu, and
  Jie Zhou.
\newblock Welm: A well-read pre-trained language model for chinese.
\newblock \emph{arXiv preprint arXiv:2209.10372}, 2022.

\bibitem[Su et~al.(2024)Su, Tian, Shen, and Cai]{su2024unraveling}
Hui Su, Zhi Tian, Xiaoyu Shen, and Xunliang Cai.
\newblock Unraveling the mystery of scaling laws: Part i.
\newblock \emph{arXiv preprint arXiv:2403.06563}, 2024.

\bibitem[Weber et~al.(2024)Weber, Fu, Anthony, Oren, Adams, Alexandrov, Lyu,
  Nguyen, Yao, Adams, et~al.]{weber2024redpajama}
Maurice Weber, Dan Fu, Quentin Anthony, Yonatan Oren, Shane Adams, Anton
  Alexandrov, Xiaozhong Lyu, Huu Nguyen, Xiaozhe Yao, Virginia Adams, et~al.
\newblock Redpajama: an open dataset for training large language models.
\newblock \emph{Advances in neural information processing systems},
  37:\penalty0 116462--116492, 2024.

\bibitem[Wu et~al.(2025)Wu, Ke, Zhou, Sun, and Ji]{roe}
Qiong Wu, Zhaoxi Ke, Yiyi Zhou, Xiaoshuai Sun, and Rongrong Ji.
\newblock Routing experts: Learning to route dynamic experts in existing
  multi-modal large language models.
\newblock In \emph{The Thirteenth International Conference on Learning
  Representations}, 2025.
\newblock URL \url{https://openreview.net/forum?id=vtT09dYPGI}.

\bibitem[Xu et~al.(2025)Xu, Wang, Luo, and
  Du]{xu2025rethinkingvisualtokenreduction}
Rui Xu, Yunke Wang, Yong Luo, and Bo~Du.
\newblock Rethinking visual token reduction in lvlms under cross-modal
  misalignment, 2025.
\newblock URL \url{https://arxiv.org/abs/2506.22283}.

\bibitem[Zellers et~al.(2019)Zellers, Holtzman, Bisk, Farhadi, and
  Choi]{zellers2019hellaswag}
Rowan Zellers, Ari Holtzman, Yonatan Bisk, Ali Farhadi, and Yejin Choi.
\newblock Hellaswag: Can a machine really finish your sentence?
\newblock \emph{arXiv preprint arXiv:1905.07830}, 2019.

\bibitem[Zhao et~al.(2025)Zhao, Ye, Fan, Tong, Xiong, Fei, Su, and
  Shen]{zhao2025skipgpt}
Anhao Zhao, Fanghua Ye, Yingqi Fan, Junlong Tong, Jing Xiong, Zhiwei Fei, Hui
  Su, and Xiaoyu Shen.
\newblock Skip{GPT}: Each token is one of a kind.
\newblock In \emph{Forty-second International Conference on Machine Learning},
  2025.
\newblock URL \url{https://openreview.net/forum?id=d7v2iUSa9s}.

\bibitem[Zheng et~al.(2025)Zheng, Koh, Ju, Nguyen, May, Webb, and
  Pan]{Zheng2025}
Yizhen Zheng, Huan~Yee Koh, Jiaxin Ju, Anh T.~N. Nguyen, Lauren~T. May,
  Geoffrey~I. Webb, and Shirui Pan.
\newblock Large language models for scientific discovery in molecular property
  prediction.
\newblock \emph{Nature Machine Intelligence}, 7\penalty0 (3):\penalty0
  437--447, 2025.
\newblock \doi{10.1038/s42256-025-00994-z}.

\end{thebibliography}
\bibliographystyle{iclr2026_conference}

\appendix
\section{Appendix}
\subsection{Statement on Large Language Model Usage}
The authors use Deepseek-r1\citep{deepseekai2025deepseekr1incentivizingreasoningcapability} solely for text refinement, including grammar checking, polishing, and condensing sections to meet length constraints. 

\subsection{Experiment Details}
\label{sec:experiment_detail}
\paragraph{Training}
Hyper-parameters differ across stages, all stages adopt the same AdamW optimizer \citep{loshchilov2017decoupled} ($\beta_1=0.9$, $\beta_2=0.95$).

\begin{itemize}
    \item \textit{Forecaster Initialization.} Constant learning rate ($1e^{-3}$), training steps (2000), batchsize (8).
    \item \textit{Router Tuning:} Constant learning rate ($2\times10^{-3}$) , training steps (2000), batchsize (16).
    \item \textit{LoRA Tuning:} Cosine annealing learning rate ($2\times10^{-3}$), training steps (2000) with warmup steps (200), batchsize (16).
\end{itemize}

It is worth emphasizing that in our experiments, we found that computing the router and LFF after normalization (e.g., RMSNorm in llama) can improve training stability—especially when the LFF involves some non-linear activations (e.g., Swish). Therefore, we strongly recommend computing the router and LFF after normalization, which is exactly the approach we adopted in our experiments (to both SkipGPT and LFF).

All experiments were conducted on a single NVIDIA RTX 6000 GPU with 48 GB VRAM. The time consumption of the three experimental phases is summarized in Table \ref{tab:time_consumption}.

\begin{table}[htbp]
    \centering
    \small
    \begin{tabular}{lc}
        \toprule
        Experimental Phase & Time Consumption \\
        \midrule
        LFF Initialization & 5 minutes \\
        Router Training & 3 hours \\
        LoRA Finetuning & 4 hours \\
        \bottomrule
    \end{tabular}
    \caption{Time consumption of different experimental phases.}
    \label{tab:time_consumption}
\end{table}

\paragraph{Evaluating}
We use lm-eval~\citep{eval-harness} for all evaluation tasks with version 0.4.9. And followed SkipGPT~\citep{zhao2025skipgpt}, tasks are evaluated with different few-shot contexts, details are listed in Table~\ref{tab:task_configs}.
\begin{table}[htbp]
    \centering
    \begin{tabular}{lcc}
        \toprule
        Task Name & Number of Few-shot Examples & Evaluation Metric \\
        \midrule
        OpenBookQA & 0 & acc\_norm \\
        Winogrande & 5 & acc \\
        PIQA & 0 & acc \\
        HellaSwag & 10 & acc\_norm \\
        BoolQ & 0 & acc \\
        ARC-Easy & 25 & acc\_norm \\
        ARC-Challenge & 25 & acc\_norm \\
        WikiText2 & 0 & word\_perplexity\\
        \bottomrule
    \end{tabular}
    \caption{Configuration of few-shot examples and evaluation metrics for different tasks.}
    \label{tab:task_configs}
\end{table}

\subsection{Comparison Methods}
\label{sec:comparison_method}
To provide a comprehensive evaluation, the proposed method is compared with several state-of-the-art approaches in static model compression and dynamic computation allocation. 
\begin{itemize}
  \item \textbf{SliceGPT} \citep{ashkboos2024slicegpt}: This method applies Principal Component Analysis (PCA) on orthogonally transformed parameters to remove entire rows and columns, achieving static parameter pruning. It results in a uniformly smaller model by permanently removing fixed structural components.
  \item \textbf{Shortened-llama} \citep{kim2024shortened}: This approach focuses on depth pruning by removing consecutive layers in LLMs to create a smaller model. It demonstrates that reducing model depth can be an efficient strategy for LLM inference.
  \item \textbf{ShortGPT} \citep{men2024shortgptlayerslargelanguage}: Leveraging Block Influence (BI), ShortGPT quantitatively estimates the importance of layers to prune less critical ones. This method is a static layer-pruning technique that aims to reduce model capacity.
  \item \textbf{Mixture-of-Depths (MoD)} \citep{raposo2024mixture}: MoD is a dynamic computation allocation method that enforces a fixed sparsity ratio per layer block. It employs a greedy routing paradigm where routers decide to execute or skip computational units for tokens.
  \item \textbf{D-LLM} \citep{jiang2024d}: This method introduces global adaptive sparsity, dynamically allocating computation across layers based on input characteristics. It refines dynamic computation by allowing more flexible computation paths across the model.
  \item \textbf{SkipGPT} \citep{zhao2025skipgpt}: A prominent dynamic computation allocation baseline, SkipGPT further refines granularity by decoupling attention and MLP operations within each layer. It uses separate routers to independently skip sub-modules, operating under the greedy routing paradigm.
\end{itemize}

\subsection{Analysis on Key-Value Cache Reduction}
\label{subsec:kvcache_analysis}
While the primary design of SkipGPT and our method focuses on computational reduction, the memory footprint of the Key-Value (KV) Cache remains a critical bottleneck in autoregressive transformer inference.  To directly target KV cache reduction, we adopt an aggressive strategy following \citep{jiang2024d}: an additional masking mechanism is applied to prevent normal tokens from attending to any skipped (or route to LFF branch) tokens in the sequence, to simulate that removing the selected tokens' key-value pairs from the cache.

The performance impact of this operation is non-negligible, as it alters the model's fundamental attention pattern. Our experiments (TABLE \ref{tab:kvr}) confirm that enforcing this strict KV cache reduction leads to a predictable degradation in model quality. The training convergence loss increases by 0.19, and the perplexity on WikiText2 rises by 2.3 points compared to the standard pruning setup which retains the full cache. This decline underscores a direct trade-off between memory compression and model fidelity; removing information from the attention context inevitably impairs the model's representational capacity.
\begin{table}[h]
\centering
\caption{Performance of models with/without KV reduction at 25\% sparsity}
\label{tab:kvr} 
\tiny
\begin{tabular}{lcc} 
\toprule
\textbf{Method} & \textbf{Validation Loss} & \textbf{Final PPL $\downarrow$} \\ 
\midrule
w.o KVR & 2.30 & 8.93 \\
w. KVR & 2.49 & 11.21 \\
\bottomrule
\end{tabular}
\end{table}

\subsection{The Selection of Lightweight Feature Forecaster}
\label{subsec:lff_analysis}

\begin{table}[h]
\centering
\caption{Performance of models with LFFs of different intermediate dimensions at 25\% sparsity. Performance saturates beyond a dimension of 50.}
\label{tab:lff_dim}
\tiny
\begin{tabular}{lccccc}  
\toprule
\textbf{Inter. Dim} & \textbf{10} & \textbf{50} & \textbf{100} & \textbf{500} & \textbf{1000} \\  
\midrule
Params. (M: $10^6$) & $\sim$0.082 & $\sim$0.41 & $\sim$0.82 & $\sim$4.10 & $\sim$8.20 \\  
Final PPL $\downarrow$ & 15.9 & 9.3 & 8.9 & 8.9 & 8.8 \\  
\bottomrule
\end{tabular}
\end{table}

The design of the LFF is central to our \textit{informed routing} paradigm. We use a two-layer linear network motivated by two factors: \textbf{first}, its simplicity helps validate our core hypothesis—that forecasting before routing stabilizes the training process—without obscuring the gains; \textbf{second}, it is highly parameter-efficient, avoiding computational overhead that could undermine acceleration.
An ablation on the LFF's intermediate dimension (Table~\ref{tab:lff_dim}) shows performance saturates beyond a dimension of 50, suggesting only a limited subset of token transformations are simple enough to be captured linearly.

Naturally, a more complex forecaster (e.g., an architecture identical to the original model block represents the theoretical upper bound of performance) could achieve better accuracy but offers less speedup, defeating the purpose of inference acceleration. Thus, the LFF lies on a Pareto frontier between performance and efficiency. Our framework allows users to configure its complexity based on specific accuracy-speed trade-offs, ensuring adaptability across scenarios.

\end{document}